\documentclass[11pt]{article}
\usepackage[final]{acl}
\usepackage{times}
\usepackage{latexsym}
\usepackage[T1]{fontenc}
\usepackage[utf8]{inputenc}
\usepackage{microtype}
\usepackage{inconsolata}
\usepackage{amsmath,amssymb}
\usepackage{booktabs}
\usepackage{tabularx}
\usepackage{graphicx}
\usepackage{float}
\usepackage{fvextra}
\usepackage{fontawesome5}

\title{We Are What We Remember: Mental-State-Driven Memory\\
for Long-Term Character Dialogue}
\author{{\large\bfseries
Zhicheng Jia\textsuperscript{1,2*},
Shuchen Wu\textsuperscript{1*\textdagger},
Zhuoshi Tong\textsuperscript{1,3*},}\\
{\large\bfseries
Haisen Liao\textsuperscript{1},
Sicheng Zhou\textsuperscript{1,4},
Jianhua Liao\textsuperscript{1},
Chao Yang\textsuperscript{1}}\\
{\normalfont
\textsuperscript{1}Livo Team, Kuaikan World\quad
\textsuperscript{2}China Agricultural University}\\
{\normalfont
\textsuperscript{3}University of Chinese Academy of Sciences\quad
\textsuperscript{4}Nanyang Technological University}\\
{\normalfont\small \textsuperscript{*}Equal contribution.\quad
\textsuperscript{\textdagger} Corresponding author:
\href{mailto:wushuchen@kkworld.com}{\nolinkurl{wushuchen@kkworld.com}}}}

\begin{document}
\maketitle
\raggedbottom
\setcounter{dbltopnumber}{1}

\begin{abstract}
Shared history in human dialogue is not merely a repository of facts, but the foundation of evolving relationships and mutual understanding. However, mainstream conversational memory systems rely on passive semantic similarity over external fact stores, frequently retrieving topically similar records while missing relationally pivotal experiences. Inspired by cognitive psychology's Self-Memory System (SMS), we propose \textbf{MindMem}, a mental-state-driven memory framework that asynchronously consolidates shared experiences into source-linked mental states of the character, the user, and their relationship. At each turn, a specialized mental-state reasoner (\textit{Mind}) contextualizes these persistent states to anticipate latent user needs, form proactive intentions, and recover supporting episodic evidence that surface-level matching overlooks. To evaluate history-sensitive adaptation beyond factual recall, we introduce \textbf{MemReal-Bench}, a longitudinal benchmark grounded in month-long real-user interactions across dyadic and group scenes. By pairing Direct Question Answering (DQA) with Situated Memory Realization (SMR), it tests whether agents bridge the gap between \textit{knowing} past facts and naturally \textit{using} shared history, world rules, and relational trajectories. On \textbf{MemReal-Bench}, \textbf{MindMem} reaches 59.03\% SMR and 82.58\% DQA, exceeding the strongest memory baseline by 24.19 and 9.20 percentage points, respectively. Direct gold-memory injection yields 50.65\% SMR. Code and dataset are available at \href{https://github.com/kkmh-org/MindMem}{\mbox{\textcolor{black}{\faGithub}\,MindMem}}.
\end{abstract}

\section{Introduction}

\begin{quote}
\begin{minipage}{\linewidth}
\itshape
\textquotedblleft{}We are who we are in good measure because of what we have learned and what we remember.\textquotedblright{}

\upshape\raggedleft\footnotesize
\mbox{--- Eric Kandel, Nobel Lecture \citep{kandel2000}}
\end{minipage}
\end{quote}

\begin{figure*}[t]
\centering
\includegraphics[width=\textwidth]{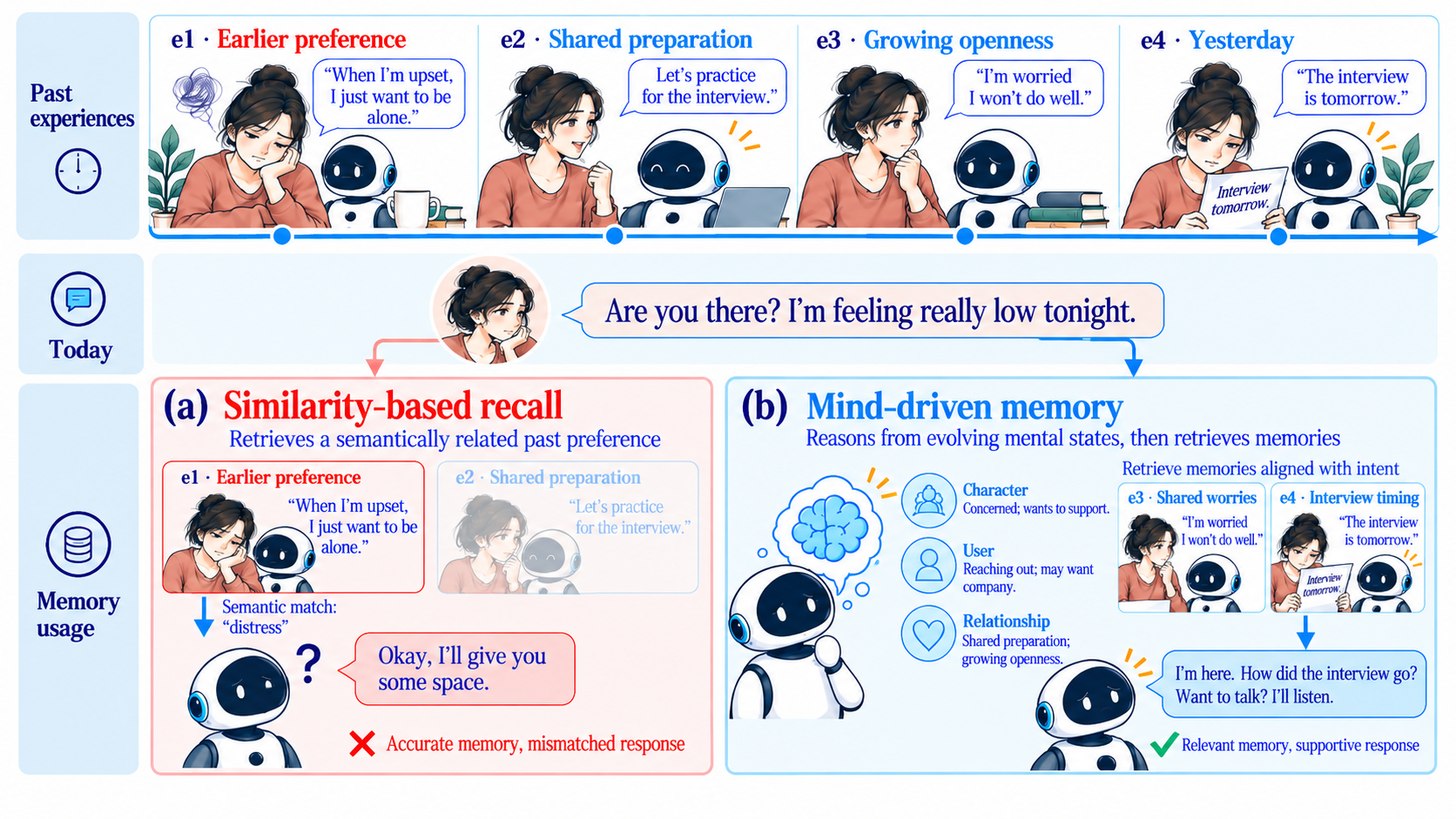}
\caption{An illustrative comparison under identical dialogue histories and current input. Passive similarity matching favors an earlier solitude preference, whereas \textbf{MindMem} reasons over evolving relational states to anticipate user needs, recover source-linked interview episodes, and offer a tentative, supportive check-in.}
\label{fig:intro-need-aligned}
\end{figure*}

Consider a scenario where a user initially expressed a preference for solitude when upset, but recently shared anxiety about an upcoming interview (Figure~\ref{fig:intro-need-aligned}). When the user reaches out feeling low today, a system relying on passive similarity matching surfaces the earlier solitude preference and inappropriately withdraws. In contrast, drawing on recent shared preparation and growing mutual trust, an ideal companion infers the underlying cause, anticipates a present need for support, and gently checks in on the interview. This contrast highlights a crucial gap in current conversational agents: the value of long-term memory lies not merely in retaining an external store of past facts, but in reasoning over shared experiences to anticipate what the partner needs in the present interaction.

Achieving such relational sensitivity remains challenging for existing architectures. Prior systems support long-term interaction through dialogue memories \citep{xu2022beyond}, recursive summaries and user profiles \citep{wang2023recursively,zhong2024memorybank}, reflection and planning \citep{park2023generative}, or model-controlled access to memory tiers \citep{packer2023memgpt}. These approaches already extend beyond passive similarity matching. Our focus is on how a persistent understanding of the user, the character, and their relationship can guide turn-level need anticipation and source-linked recall: why a past experience matters now and what conversational direction it should support.

To address this, we draw inspiration from the Self-Memory System (SMS) in cognitive psychology \citep{conway2000}, where a dynamic \textit{working self} maintains active goals to modulate access to an \textit{autobiographical knowledge base}. Translating this mechanism to conversational agents, we propose \textbf{MindMem}, a mental-state-driven memory framework. In our approach, shared experiences are asynchronously consolidated into source-linked mental states of the character, the user, and their relationship. Functioning as the working self, a specialized reasoner (\textit{Mind}) contextualizes these persistent states within the ongoing dialogue to hypothesize latent user needs, formulate proactive intentions, and trace source links to recover supporting episodes that surface-level matching misses. A response generator then integrates this mental-state interpretation with the selected memories to craft replies aligned with both immediate needs and the long-term relational trajectory.

Validating such history-sensitive adaptation requires evaluating whether models bridge the gap between \textit{knowing} isolated facts and \textit{using} implicit relational dynamics. While existing benchmarks assess long-context recall \citep{maharana2024,wu2025} or cue-based memory utilization \citep{sumida2026,he2025,wu2026}, we focus on the joint continuity of character identity, relational context, and proactive response choices. To this end, we introduce \textbf{MemReal-Bench}, a longitudinal benchmark seeded from month-long real-user interactions across dyadic and group scenes. Pairing Direct Question Answering (DQA) with Situated Memory Realization (SMR), it tests whether agents naturally enact world rules, character principles, user preferences, and relational trajectories under implicit triggers.

Our main contributions are threefold:
\begin{itemize}
    \item \textbf{Mental-State-Driven Framework:} We propose \textbf{MindMem}, coupling asynchronous state consolidation with a trainable reasoner (\textit{Mind}) for need anticipation and source-linked memory retrieval.
    \item \textbf{Real-Interaction-Grounded Benchmark:} We introduce \textbf{MemReal-Bench} to jointly evaluate explicit factual recall (DQA) and situated memory realization (SMR) across longitudinal dyadic and group interactions.
    \item \textbf{Knowing-versus-Using Insights:} \textbf{MindMem} improves SMR by 24.19 percentage points and DQA by 9.20 points over the strongest memory baseline. Its SMR also exceeds that of direct gold-memory injection, illustrating the gap between access to target memories and their situated use.
\end{itemize}

\section{Related Work}

\subsection{Long-Term Dialogue Agents}

Long-term dialogue agents leverage accumulated interactions to maintain continuity and personalize responses, early on through recursive summaries \citep{wang2023recursively,zhong2024memorybank} or reflective abstractions that support planning \citep{park2023generative}. Recent work advances this paradigm along two complementary axes: memory organization and retrieval control. To enrich organization, A-Mem links atomic notes into evolving knowledge networks \citep{xu2025amem}; to refine control, Self-RAG integrates adaptive retrieval with self-reflection \citep{asai2024selfrag}. In multi-session dialogue, MemGuide aligns memory selection with task intent and missing information \citep{du2026}, while MapDia couples historical-topic retrieval with policies for reintroducing past topics \citep{wu2025mapdia}. Together, these methods connect memory structure with deliberate decisions about \textit{when} and \textit{why} to access past experiences.

In character-based interaction, early frameworks such as Character-LLM \citep{shao2023characterllm} and RoleLLM \citep{wang2024rolellm} focus on role profiles and role-consistent generation. Long-term character dialogue additionally requires interpreting shared history through an evolving understanding of the partner. DualMem separates factual records from persona-conditioned insights \citep{zhang2026}. ToMAgent trains explicit Theory-of-Mind reasoning with dialogue lookahead to support goal-directed social interaction \citep{hwang2026}, while PsyProbe combines structured user-state modeling, memory construction, and strategy planning for exploratory counseling \citep{park2026psyprobe}. These systems provide complementary mechanisms for persona interpretation, social reasoning, and proactive dialogue. In this context, \textbf{MindMem} studies their integration with a persistent, source-linked store of character, user, and relationship states: background consolidation revises these states across sessions, and a specialized turn-level reasoner uses them to anticipate needs and select supporting episodes.

\subsection{Memory and Dialogue Evaluation}

Long-term memory benchmarks evaluate conversational agents at progressively deeper levels of granularity. Foundational benchmarks focus on factual access and cross-session reasoning, such as multimodal QA and event summarization in LoCoMo \citep{maharana2024}, or temporal updates and abstention in LongMemEval \citep{wu2025}. Moving beyond passive factual extraction, recent benchmarks examine how memory shapes dialogue generation and constraint adherence. RoleMemo \citep{zhang2026} and LoCoMo-Plus \citep{li2026locomoplus} evaluate persona-conditioned interpretation of historical evidence and adherence to implicit user-state, goal, and value constraints. At the response level, MemUse \citep{sumida2026}, MADial-Bench \citep{he2025}, and StratMem-Bench \citep{wu2026} assess naturally cued memory integration, emotion-cued activation and intimacy, and strategic memory selection in character dialogue, respectively.

Advancing this shift from knowing to situated using, \textbf{MemReal-Bench} pairs explicit factual verification (DQA) with Situated Memory Realization (SMR) across longitudinal dyadic and group interactions. Crucially, it evaluates how shared experience informs subtle, proactive response choices---such as conversational initiative, world-rule adherence, and interpersonal distance---under natural triggers that do not require artificial or explicit historical recounting.

\section{Mental-State-Driven Memory Framework}

\begin{figure*}[t]
\centering
\includegraphics[width=\textwidth]{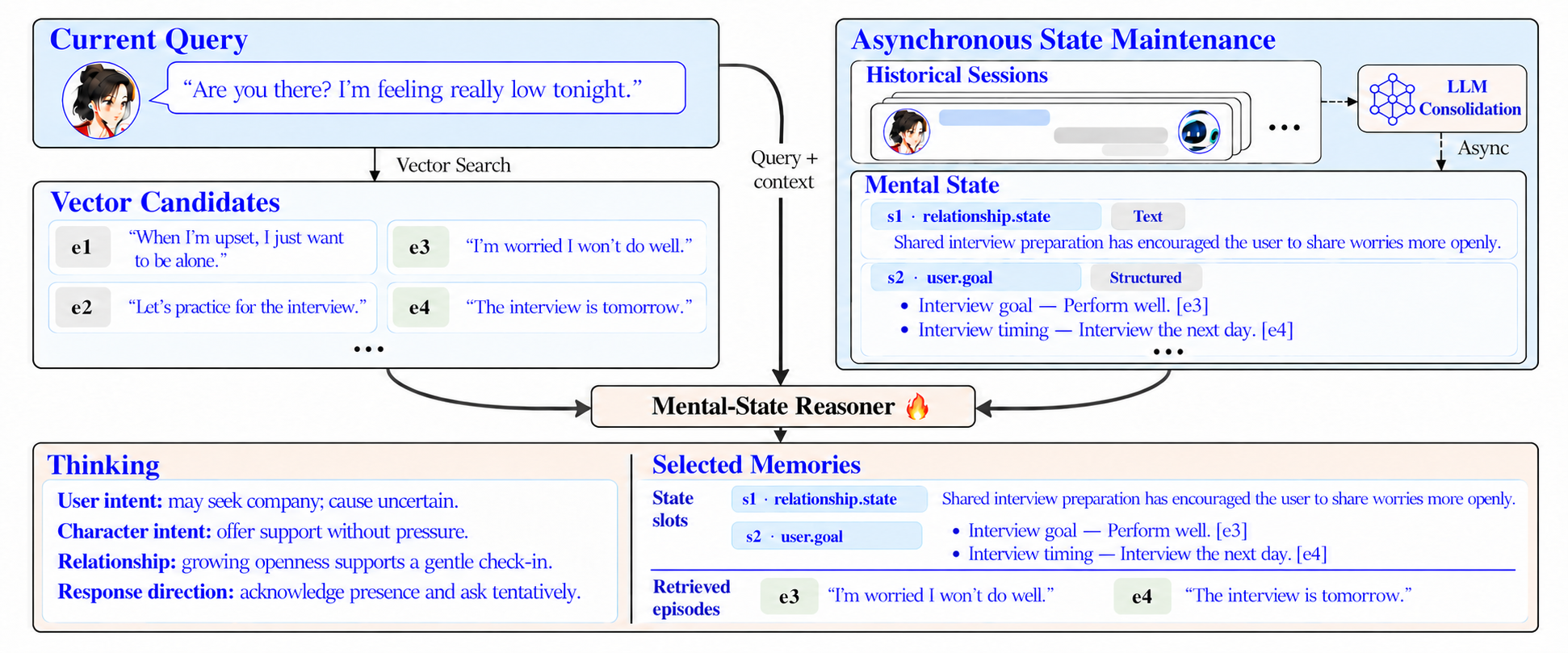}
\caption{Overview of \textbf{MindMem}. Asynchronous consolidation maintains source-linked state slots across 40 dimensions. At query time, given current context, persistent states, and vector candidates, \textit{Mind} reasons over the interaction to anticipate user needs and select memories for source-linked retrieval. Dashed arrows denote asynchronous updates; the flame marks the trainable reasoner.}
\label{fig:framework}
\end{figure*}

Inspired by goal-directed memory access in the Self-Memory System \citep{conway2000}, \textbf{MindMem} decouples background state consolidation from online mental-state reasoning (Figure~\ref{fig:framework}). Across long-term interactions, the framework maintains raw dialogue spans in an episodic store $\mathcal{E}_t$ alongside source-linked mental states in $\mathcal{M}_t$. Given a user utterance $u_t$, local scene context $c_t$, and character specification $p$ at turn $t$, the turn-level reasoner \textit{Mind} contextualizes $\mathcal{M}_t$ to guide evidence retrieval and generate a history-consistent response $y_t$.

\subsection{Asynchronous State Consolidation}

An LLM consolidator $C$ asynchronously revises $\mathcal{M}_t$ from completed interactions, retaining provenance pointers to the underlying dialogue spans in $\mathcal{E}_t$ \citep{park2023generative}. In \textbf{MemReal-Bench}, $C$ uses \texttt{deepseek-v4-flash}. An update is triggered when 10 new memory records have accumulated, using a recent context window of 20 dialogue turns. Online reasoning reads the latest committed snapshot alongside recent dialogue, which may contain observations not yet consolidated. Immediate needs and response intentions are inferred at query time rather than written back as new historical evidence.

The schema contains 40 dimensions across seven families, grouped into participant states, dyadic dynamics, and shared context (Appendix~\ref{app:slot-schema}). Dimensions use either narrative text or structured records to represent character and user dispositions, relationship states, and shared experiences. Consolidation revises prior interpretations while retaining their evidence and situational scope: a desire for company during interview preparation can coexist with a general preference for solitude. Temporal cutoffs and character visibility constrain accessible records; attribution and grounding instructions distinguish the real user from a fictional avatar and observations from tentative interpretations (Appendix~\ref{app:information-boundaries}).

\subsection{Mental-State-Guided Recall and Generation}
\label{sec:active-recall}

Dense retrieval first uses $(u_t,c_t)$ to obtain top-$K$ episodic candidates $\mathcal{C}_t$ from $\mathcal{E}_t$ \citep{karpukhin2020dpr}. We set $K=20$ for \textbf{MemReal-Bench}. In a single model call, \textit{Mind}, parameterized by $\pi_\theta$, reads $(u_t,c_t,p)$, the accessible contents of all 40 state dimensions, and $\mathcal{C}_t$, and produces
\[
z_t=(\tau_t,\mathcal{S}_t,\mathcal{V}_t).
\]
The \textit{Thinking} trace $\tau_t$ interprets current needs, character constraints, and relational context to form a tentative response intention. It guides selection of state entries $\mathcal{S}_t\subseteq\operatorname{Slots}(\mathcal{M}_t)$ and episodic candidates $\mathcal{V}_t\subseteq\mathcal{C}_t$; either selection may be empty (Appendix~\ref{app:mind-action-space}).

State selection provides an access path beyond the initial vector candidates. For an indirect utterance such as ``I feel low today,'' \textit{Mind} may select an ongoing interview goal whose source episodes have little lexical similarity to the utterance. The retrieval pipeline resolves the selected entries' provenance links and assembles the evidence:
\[
\begin{aligned}
\mathcal{L}_t&=\operatorname{LinkedEvidence}(\mathcal{S}_t;\mathcal{E}_t),\\
\mathcal{D}_t&=\operatorname{Assemble}(\mathcal{V}_t\cup\mathcal{L}_t).
\end{aligned}
\]
Link resolution applies temporal and visibility cutoffs; assembly deduplicates and orders the episodes within the context budget. Thus, supporting episodes can enter $\mathcal{D}_t$ even when they are absent from $\mathcal{C}_t$.

The response generator $G_\phi$ conditions on $(u_t,c_t,p)$, $\tau_t$, selected states $\mathcal{S}_t$, and assembled episodes $\mathcal{D}_t$ to produce $y_t$ \citep{lewis2020rag}. Provenance and temporal tags connect state interpretations to their supporting episodes. The generator is instructed to check tentative intentions against this evidence and explicit user requests, revising unsupported assumptions when needed \citep{gao2023alce}. In Figure~\ref{fig:framework}, the interview state exposes anxiety ($e_3$) and timing ($e_4$), supporting a gentle check-in.

\subsection{Training the Mental-State Reasoner}
\label{sec:training-mind}

Training adapts only the compact reasoner $\pi_\theta$ while keeping consolidation, dense retrieval, and the generator $G_\phi$ frozen.

\paragraph{Training examples and supervision.}
Dropping turn indices, each training input is $x=(u,c,p,\mathcal{M},\mathcal{C})$: a current utterance and local context, the character specification, accessible state entries, and pre-retrieved episodic candidates. We pair these inputs with reference decisions $z^*=(\tau^*,\mathcal{S}^*,\mathcal{V}^*)$ curated over training histories. The reference \textit{Thinking} trace $\tau^*$ interprets the interaction and guides memory use; $\mathcal{S}^*$ and $\mathcal{V}^*$ identify state entries and candidate episodes in the supplied input. Reference \textit{Thinking} traces and memory selections are generated by Claude Opus 5.5 and manually verified. Examples include empty selections when history is unneeded. Source-linked evidence is resolved from selected states by the retrieval pipeline. Data sources, scale, and partitioning are reported in Section~\ref{sec:experimental-setup}, with the training procedure in Appendix~\ref{app:training-procedure}.

\paragraph{Supervised initialization and reinforcement learning.}
We initialize $\pi_\theta$ via supervised fine-tuning (SFT), minimizing the negative log-likelihood of $z^*$ conditioned on $x$ \citep{ouyang2022instructgpt}.

To align selection with downstream memory use, we further refine $\pi_\theta$ via Group Relative Policy Optimization (GRPO; \citealp{shao2024deepseekmath}). For each input $x$, we sample a group of $G$ decisions $\{z^{(i)}\}_{i=1}^G$, assemble their source-linked evidence, and generate corresponding replies $\{y^{(i)}\}_{i=1}^G$ using the frozen generator $G_\phi$. Each pair $(z,y)$ receives a composite reward $r(x,z,y)\in[0,1]$:
\[
\begin{aligned}
r(x,z,y)=g(z)\bigl[&\lambda_{\mathrm{fmt}}+\lambda_{\mathrm{acc}}a(x,z)\\
                   &+\lambda_{\mathrm{qual}}q(x,y)\bigr],
\end{aligned}
\]
where the nonnegative weights sum to one. The format gate $g(z)\in\{0,1\}$ equals $1$ only if $z$ is structurally valid and references admissible IDs in $\mathcal{M}$ and $\mathcal{C}$. Selection accuracy $a(x,z)$ averages the set-level $F_1$ scores of $\mathcal{S}$ and $\mathcal{V}$ against reference targets, while an LLM judge scores downstream response quality $q(x,y)\in[0,1]$ for historical grounding, character continuity, relational consistency, and calibrated uncertainty (Appendix~\ref{app:reward-spec}). GRPO updates $\pi_\theta$ using group-normalized advantages, clipped policy ratios, and a KL penalty toward the SFT reference, optimizing memory selection through downstream reply quality without backpropagating through retrieval or generation.

\section{Experiments}
\label{sec:experiments}

\subsection{MemReal-Bench}
\label{sec:benchmark}

\begin{table*}[t]
\centering\footnotesize
\setlength{\tabcolsep}{5pt}
\renewcommand{\arraystretch}{1.12}
\begin{tabular*}{\textwidth}{@{\extracolsep{\fill}}l l l *{6}{c}@{}}
\toprule
& \multicolumn{2}{c}{\textbf{Dialogue setting}} & \multicolumn{6}{c}{\textbf{Evaluation coverage}}\\
\cmidrule(lr){2-3}\cmidrule(l){4-9}
\textbf{Benchmark} & \textbf{Interaction} & \textbf{Structure} & DQA & FR & WC & RR & RoR & PI\\\midrule
LoCoMo & Dyadic & Multi-session & $\checkmark$ & $\times$ & $\times$ & $\times$ & $\times$ & $\times$\\
LongMemEval & Dyadic & Long-term & $\checkmark$ & $\times$ & $\times$ & $\times$ & $\times$ & $\times$\\
MemUse & Dyadic & Long-term & $\checkmark$ & $\checkmark$ & $\times$ & $\times$ & $\times$ & $\times$\\
EverMemBench & Group & Long-term & $\checkmark$ & $\times$ & $\times$ & $\times$ & $\times$ & $\checkmark$\\
\textbf{MemReal-Bench} & Dyadic + group & Long-term & $\checkmark$ & $\checkmark$ & $\checkmark$ & $\checkmark$ & $\checkmark$ & $\checkmark$\\\bottomrule
\end{tabular*}
\caption{Comparison with prior conversational memory benchmarks: LoCoMo \citep{maharana2024}, LongMemEval \citep{wu2025}, MemUse \citep{sumida2026}, and EverMemBench \citep{hu2026evermembench}. $\checkmark$ / $\times$: coverage / no coverage. DQA and SMR dimensions are defined in Section~\ref{sec:benchmark-metrics}.}
\label{tab:benchmark-comparison}
\end{table*}

\paragraph{Construction.}
\label{sec:benchmark-construction}
\textbf{MemReal-Bench} jointly evaluates factual recall and situated memory realization across long-term dyadic and group character dialogue. Its seeds are distilled from a month-long source collection of real-user and character interactions; for user privacy, raw chat logs are never passed to the generator or included in the benchmark. A three-stage pipeline (Appendix~\ref{app:benchmark-construction}) (1) distills user personas, interaction patterns, and relationship trajectories; (2) synthesizes multi-session dialogue mixing dyadic and group scenes, annotated with participants, session/turn coordinates, and visibility scopes; and (3) extracts atomic memory records from the realized transcripts to build natural-response triggers, target-memory rubrics, and Direct Question Answering (DQA) items. The evaluated benchmark comprises 310 Situated Memory Realization (SMR) probes and 511 scored DQA questions over 4{,}617 dialogue turns. As summarized in Table~\ref{tab:benchmark-comparison}, it evaluates not only factual recall but whether responses naturally respect world rules, relational trajectories, character principles, and user preferences.

\paragraph{Evaluation Protocol.}
\label{sec:benchmark-inputs}
For probe $i$, the task input is $x_i=(H_i^{\leq b_i,r_i},c_i,u_i,p_i)$, where $H_i^{\leq b_i,r_i}$ is the historical prefix visible to responding character $r_i$ at cutoff $b_i$, $c_i$ is the scene context, $u_i$ is the current user utterance, and $p_i$ is the character specification. The \textit{GroundTruth} condition additionally receives annotated target memories. Scoring rubrics are reserved for the judge. Instances enforce temporal validity over superseded states \citep{wu2025}, visibility boundaries across group and dyadic scenes \citep{kim2023}, and separation between the real user and the fictional avatar; natural triggers cue the situation without leaking the target \citep{li2026locomoplus,sumida2026}.

\begin{figure*}[t]
\centering
\includegraphics[width=\textwidth]{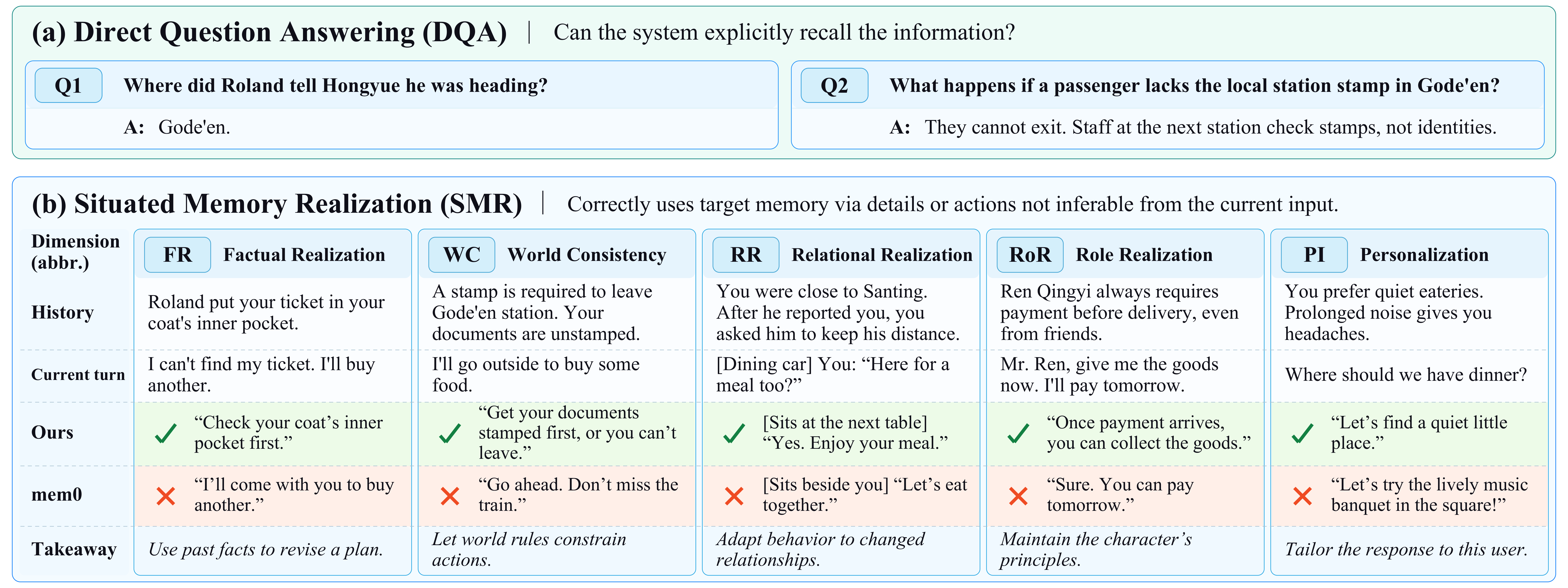}
\caption{Illustrative examples of knowing versus using memory. DQA tests explicit factual recall, whereas SMR evaluates whether responses naturally enact history-specific details or actions under natural triggers across FR, WC, RR, RoR, and PI.}
\label{fig:smr-instance}
\end{figure*}

\paragraph{Metrics: DQA and SMR.}
\label{sec:benchmark-metrics}
\textbf{DQA} tests explicit factual access via binary-scored questions judged against annotated answer points; abstentions count as incorrect. \textbf{SMR} tests whether a response realizes the target memory through history-specific details or actions not inferable from $u_i$ alone. Recounting the past is unnecessary: keeping a requested distance or honoring an established principle can realize memory, whereas trigger echoing, generic empathy, or state contradictions cannot. SMR probes span five dimensions (Figure~\ref{fig:smr-instance}; Table~\ref{tab:scoring}): \textit{Factual Realization} (FR), \textit{World Consistency} (WC), \textit{Relational Realization} (RR), \textit{Role Realization} (RoR), and \textit{Personalization} (PI). An identity-blind LLM judge assigns binary verdicts, and overall SMR is their micro-average (Appendix~\ref{app:benchmark-dimensions}).

\subsection{Experimental Setup}
\label{sec:experimental-setup}

\paragraph{Training Data.}
Training examples are constructed from an internal corpus of de-identified, real-world user--agent interactions using the input and supervision format in Section~\ref{sec:training-mind}. We use 2,000 examples for SFT initialization and reuse the same inputs for GRPO. Training and benchmark test histories are partitioned by source user with no overlap, and benchmark annotations are excluded from optimization.

\paragraph{Implementation Details.}
On \textbf{MemReal-Bench}, conditions share test instances, historical cutoffs, task-specific prompt templates and decoding settings, the \texttt{deepseek-v4-pro} response generator, and the \texttt{deepseek-v4-flash} judge. The SMR template supplies the character specification, current user utterance, and last 10 dialogue turns visible to the responding character, with the recent turns omitted for \textit{NoMem}. DQA uses a separate factual-answer template with an empty scene context. Memory systems ingest visible histories chronologically up to $b_i$ and supply at most 16 evidence records to the generator \citep{xu2025amem}.

We instantiate \textit{Mind} with Qwen3.5 backbones at 0.8B and 2B and supply all 40 state dimensions. Main results and diagnostics use SFT+GRPO-2B. Before each \textbf{MemReal-Bench} probe, \textbf{MindMem} completes pending consolidation even when fewer than 10 new records have accumulated since the last consolidation. Its generator input combines deduplicated vector-selected and slot-linked evidence with selected state slots and \textit{Thinking} guidance. The cap applies to evidence records, and input token counts vary across systems. The 2B selector completes its output in a median of 1.7 seconds before response generation (Table~\ref{tab:scale}).

\paragraph{Baselines.}
We compare \textbf{MindMem} with Mem0 \citep{chhikara2025mem0}, A-Mem \citep{xu2025amem}, and SimpleMem \citep{liu2026simplemem}. We use the latest Mem0 implementation available at the time of evaluation. \textit{NoMem} omits long-term memory and recent dialogue. The gold-memory injection reference \textit{GroundTruth} supplies the annotated target memories directly to the shared generator in place of retrieved records. On LoCoMo \citep{maharana2024} and LongMemEval-S \citep{wu2025}, we follow Mem0's open-source \texttt{memory-benchmarks} protocol, with chronological ingestion, time-ordered retrieval, a fixed embedding model, and \texttt{deepseek-v4-flash} as both answerer and judge. We use 1{,}540 non-adversarial LoCoMo questions and all 500 LongMemEval-S questions (Appendix~\ref{app:experimental-config}).
\subsection{Main Results}
\label{sec:main-results}

\begin{table*}[t]
\centering\small
\setlength{\tabcolsep}{4pt}
\renewcommand{\arraystretch}{1.08}
\begin{tabular*}{\textwidth}{@{\extracolsep{\fill}}l *{9}{r}@{}}
\toprule
& \multicolumn{6}{c}{\textbf{SMR (\%)}} & \textbf{DQA (\%)} & \multicolumn{2}{c}{\textbf{Standard benchmarks (\%)}}\\
\cmidrule(lr){2-7}\cmidrule(lr){8-8}\cmidrule(l){9-10}
\textbf{Memory system} & \shortstack[r]{FR\\(143)} & \shortstack[r]{WC\\(33)} & \shortstack[r]{RR\\(47)} & \shortstack[r]{RoR\\(52)} & \shortstack[r]{PI\\(35)} & \shortstack[r]{\textbf{All}\\(310)} & \shortstack[r]{\textbf{All}\\(511)} & LoCoMo & LongMemEval-S\\\midrule
\multicolumn{10}{@{}l}{\textit{Reference settings}}\\
NoMem & 7.69 & 3.03 & 6.38 & 15.38 & 11.43 & 8.71 & 0.00 & --- & ---\\
GroundTruth & 53.85 & 63.64 & 42.55 & 61.54 & 20.00 & 50.65 & 96.67 & --- & ---\\\midrule
\multicolumn{10}{@{}l}{\textit{Memory systems}}\\
Mem0 & 32.87 & 33.33 & 25.53 & 32.69 & 8.57 & 29.03 & 71.04 & 85.19 & 75.40\\
A-Mem & 34.97 & 42.42 & 27.66 & 48.08 & 17.14 & 34.84 & 73.39 & 93.18 & 71.40\\
SimpleMem & 36.36 & 39.39 & 23.40 & 44.23 & 17.14 & 33.87 & 72.80 & \textbf{93.70} & 72.20\\
\textbf{MindMem} & \textbf{61.54} & \textbf{66.67} & \textbf{55.32} & \textbf{61.54} & \textbf{42.86} & \textbf{59.03} & \textbf{82.58} & 91.95 & \textbf{87.20}\\\bottomrule
\end{tabular*}
\caption{Main results on \textbf{MemReal-Bench}, LoCoMo \citep{maharana2024}, and LongMemEval-S \citep{wu2025}. Parentheses give the number of scored items. \textit{GroundTruth} is the gold-memory injection reference. Bold marks the best memory-system score per column; ``---'' denotes unavailable results.}
\label{tab:main}
\end{table*}

\paragraph{Comparison with Memory Systems.}
\textbf{MindMem} outperforms all memory baselines on DQA and every SMR dimension (Table~\ref{tab:main}). Gains over the strongest baseline per dimension are largest for relational realization (RR, +27.66 points) and personalization (PI, +25.71). Against A-Mem, the strongest baseline on overall SMR and DQA, a paired permutation analysis grouped by cutoff session gives $p=4.66\times10^{-10}$ for SMR and $p=0.00650$ for DQA (Appendix~\ref{app:paired-system-comparison}).

The same SFT+GRPO-2B checkpoint scores 91.95\% on LoCoMo, behind SimpleMem and A-Mem on the non-adversarial question-answering split. It leads the evaluated systems on LongMemEval-S (87.20\%), which includes knowledge-update questions alongside other long-term recall tasks.

\paragraph{Comparison with Gold Memory.}
\textit{GroundTruth} answers 96.67\% of DQA questions. With gold target memories available, it still fails on 153 of 310 SMR probes (49.35\%), including 28 of 35 personalization probes (80.00\%). Personalization calls for identifying which established preference applies in the current situation. \textbf{MindMem} reaches 59.03\% overall SMR compared with 50.65\% under direct gold-memory injection, and matches \textit{GroundTruth} on RoR. The active-recall ablation examines the joint contribution of response intentions, selected states, and their source-linked evidence (Section~\ref{sec:ablations}).

\begin{table}[t]
\centering\small
\setlength{\tabcolsep}{3pt}
\renewcommand{\arraystretch}{1.08}
\begin{tabular*}{\columnwidth}{@{\extracolsep{\fill}}l rr rrr@{}}
\toprule
& \multicolumn{2}{c}{\textbf{Know$\rightarrow$Use}} & \multicolumn{3}{c}{\textbf{History (sessions)}}\\
\cmidrule(lr){2-3}\cmidrule(l){4-6}
\textbf{System} & Real. & Gap$\downarrow$ & 1--15 & 16--30 & 31--43\\\midrule
GroundTruth & 55.6 & 41.0 & 51.2 & 52.7 & 48.2\\\midrule
Mem0 & 32.5 & 38.1 & 32.1 & 28.6 & 27.2\\
A-Mem & 32.6 & 37.0 & 38.1 & 33.9 & 33.3\\
SimpleMem & 30.0 & 35.9 & 36.9 & 32.1 & 33.3\\
\textbf{MindMem} & \textbf{62.6} & \textbf{20.1} & \textbf{58.3} & \textbf{53.6} & \textbf{64.9}\\\bottomrule
\end{tabular*}
\caption{Diagnostics on \textbf{MemReal-Bench}. \textit{Real.}\ (\%): SMR on system-specific probes with all linked DQA answers correct. \textit{Gap} (pp): question-micro DQA minus probe-micro SMR over 176 linked FR/WC probes. \textit{History} (\%): SMR by cutoff session (the three buckets partition the same 310 SMR probes: $n=84$, $112$, and $114$ probes for sessions 1--15, 16--30, and 31--43, respectively). Definitions: Section~\ref{sec:know-use} and Appendix~\ref{app:benchmark-dimensions}. Bold marks the best memory-system rate or gap.}
\label{tab:diagnostic}
\end{table}

\paragraph{Performance Across Interaction Stages.}
Table~\ref{tab:diagnostic} (right) reports SMR across three cutoff-session buckets. \textbf{MindMem} achieves 58.3\%, 53.6\%, and 64.9\% in the early, middle, and late buckets, respectively, outperforming the memory baselines in all three interaction stages. Its margin over \textit{GroundTruth} varies across buckets, from 0.9 to 16.7 percentage points.

\subsection{Knowing versus Using}
\label{sec:know-use}

\paragraph{Paired Knowing--Using Gap.}
For DQA-linked FR/WC probes, \textit{Gap} quantifies the aggregate disparity between explicit recall and situated realization by subtracting probe-micro SMR from question-micro DQA (Table~\ref{tab:diagnostic}; formal definitions in Appendix~\ref{app:benchmark-dimensions}). Among memory systems, \textbf{MindMem} achieves the smallest Gap (20.1 pp), together with the highest DQA and SMR scores. The shared-target analysis below further examines realization when all compared memory systems correctly answer the linked DQA questions.

\paragraph{Realization on Shared Targets.}
Table~\ref{tab:diagnostic} reports \textit{Real.} on each system's own set of probes with all linked DQA questions answered correctly. To compare SMR on the same targets, we restrict the analysis to probes for which Mem0, A-Mem, SimpleMem, and \textbf{MindMem} all answer every linked DQA question correctly. On this shared subset of 36 probes, \textbf{MindMem} exceeds the strongest baseline by 19.4 percentage points. This comparison shows that explicit factual recall alone does not guarantee situated use.

\paragraph{Where Use Fails.}
We inspect the recorded generator inputs for 100 of 127 failed \textbf{MindMem} responses in a targeted, dimension-stratified audit focused on failures of situated use (Appendix~\ref{app:failure-taxonomy}). The sample contains 44 FR, 11 WC, 11 RR, 15 RoR, and 19 PI cases. In 36 audited responses, the target fails to reach the generator. In the other 64, the target reaches the generator and the response fails to realize it. These cases mirror the failures observed with direct gold-memory injection.

\subsection{Ablation Study}
\label{sec:ablations}

Using SFT+GRPO-2B as \textbf{Full}, we evaluate three variants (Table~\ref{tab:ablation}). (1) \textit{w/o GRPO} uses the SFT-2B checkpoint to assess the alignment gain of GRPO beyond supervised initialization. (2) \textit{w/o specialized Mind} uses \texttt{deepseek-v4-flash} as the reasoner under the same active-recall interface. (3) \textit{w/o active recall} removes the \textit{Mind}-driven active-recall pathway, including the response intention $\tau_t$, selected states $\mathcal{S}_t$, and source-linked evidence $\mathcal{L}_t$. The variant retains dense retrieval of up to 16 raw dialogue spans from the episodic store $\mathcal{E}_t$. All conditions share test histories, historical cutoffs, the response generator, and judge settings (Appendix~\ref{app:experimental-config}).

\textbf{Full} exceeds \textit{w/o GRPO} by 4.84 percentage points in SMR and 0.58 points in DQA. Replacing the specialized \textit{Mind} with the general-purpose model reduces these scores by 17.09 and 2.15 points, respectively. Further backbone and stage comparisons are reported in Table~\ref{tab:scale}.

The vector-only variant scores 66.34\% on DQA, below the memory baselines in Table~\ref{tab:main}. Mem0 indexes memories extracted from dialogue, and SimpleMem compresses dialogue into structured memory units. Restoring the complete active-recall pathway raises SMR by 29.35 points and DQA by 16.24 points to 82.58\%.

\begin{table*}[t]
\centering\small
\setlength{\tabcolsep}{3pt}
\renewcommand{\arraystretch}{1.08}
\begin{tabularx}{\textwidth}{@{}l >{\raggedright\arraybackslash}X *{7}{r}@{}}
\toprule
& & \multicolumn{6}{c}{\textbf{SMR (\%)}} & \textbf{DQA (\%)}\\
\cmidrule(lr){3-8}\cmidrule(l){9-9}
\textbf{Variant} & \textbf{Modification} & FR & WC & RR & RoR & PI & \textbf{All} & \textbf{All}\\\midrule
\textbf{Full} & SFT+GRPO-2B; up to 16 evidence records, selected slots, and \textit{Thinking} & \textbf{61.54} & \textbf{66.67} & \textbf{55.32} & \textbf{61.54} & \textbf{42.86} & \textbf{59.03} & \textbf{82.58}\\
w/o GRPO & Replace SFT+GRPO-2B with SFT-2B & 59.44 & 60.61 & 51.06 & 50.00 & 37.14 & 54.19 & 82.00\\
w/o specialized Mind & Use \texttt{deepseek-v4-flash} under the same active-recall interface & 44.76 & 57.58 & 40.43 & 38.46 & 22.86 & 41.94 & 80.43\\
w/o active recall & Remove the active-recall pathway. Retain up to 16 vector-retrieved dialogue spans & 32.17 & 36.36 & 27.66 & 30.77 & 14.29 & 29.68 & 66.34\\\bottomrule
\end{tabularx}
\caption{Ablation study on \textbf{MemReal-Bench} (\%). \textbf{Full} uses the SFT+GRPO-2B configuration reported in Table~\ref{tab:main}. All conditions share the same 310 SMR probes, 511 DQA questions, response generator, and judge. Bold marks the highest score in each metric column.}
\label{tab:ablation}
\end{table*}

\subsection{Human Evaluation}
\label{sec:human-eval}

We sample 90 responses across the four memory systems and two reference conditions (\textit{NoMem} and \textit{GroundTruth}) in Table~\ref{tab:main}. The sample includes 60 SMR responses (two per condition in each dimension) and 30 DQA responses (five per condition). Ten blind annotators label the responses independently. Majority vote determines the human reference, with a senior expert resolving ties. Agreement between the judge and the human reference is 80\% on SMR (Cohen's $\kappa=0.60$) and 93.3\% on DQA ($\kappa=0.86$) \citep{cohen1960}. Fleiss' $\kappa$ for inter-annotator agreement is 0.61 on SMR and 0.85 on DQA \citep{fleiss1971}.

\section{Conclusion}

We introduced \textbf{MindMem}, which consolidates shared experiences into persistent character, user, and relationship states and uses turn-level reasoning to guide source-linked episodic recall and conversational responses. We also introduced \textbf{MemReal-Bench}, a benchmark seeded by real interactions that evaluates explicit factual recall (DQA) and situated memory realization (SMR). On this benchmark, \textbf{MindMem} improves SMR by 24.19 percentage points over the strongest memory baseline. Its advantage remains 19.4 points on shared probes where all compared memory systems correctly answer every linked DQA question, revealing substantial differences in situated use even when explicit recall succeeds. Failure analysis further identifies response realization as a remaining bottleneck even when target memories reach the generator. Together, these findings support coupling memory access with reasoning about how shared history should shape the current response, and evaluating both factual recall and situated use in long-term dialogue.

\section*{Limitations}

\paragraph{State size and update delay.}
\textit{Mind} reads the full visible contents of all 40 state dimensions. Accumulated entries can increase prompt length and inference cost. Asynchronous consolidation can also leave committed snapshots behind recent observations. Local context may contain corrections absent from the snapshot, and the reliability of resolving such conflicts requires measurement.

\paragraph{Interpretation and generator dependence.}
Consolidation can overgeneralize temporary preferences, and \textit{Mind} can infer unsupported intentions or relationship changes. LLM judges can exhibit position, verbosity, and self-enhancement biases \citep{zheng2023judge}. Training against a fixed generator and reward judge may favor their response styles or selection preferences; transfer across generators has not been established.

\paragraph{Empirical scope.}
While our evaluation covers month-long dyadic and group interactions, extending validation to broader domains and multi-month lifelong settings remains an open direction for future work. Relational appropriateness depends on annotation criteria and judge calibration, and selection agreement with one reference may underrepresent valid alternatives. Larger-scale human assessment and broader longitudinal comparisons remain important directions for establishing these capabilities at scale.

\section*{Ethical Considerations}

\paragraph{Dialogue privacy and data use.}
Longitudinal character dialogue may contain sensitive disclosures and identifying information. Research use requires appropriate consent or other authorization, suitable de-identification, and any applicable ethics review. Preparation should minimize retained personal information while preserving consistent pseudonymous entities within each history. Release decisions must address re-identification through linked events and avoid exposing raw personal conversations or source links. All benchmark trajectories in \textbf{MemReal-Bench} are synthesized from distilled persona and relational patterns; original private chat logs are neither supplied to the dialogue generator nor included as benchmark histories. Distilled seeds and generated outputs require screening for sensitive or identifying attributes before release.

\paragraph{Interaction boundaries.}
Inferring user needs and adapting interpersonal distance can introduce unwanted intimacy, anthropomorphism, or emotional dependence. The framework is designed to distinguish observations from tentative interpretations, separate real-user and avatar information, and respect character visibility and current explicit requests. Deployment should avoid presenting character inferences as diagnosis or professional support, and should provide ways to correct or remove retained information.

\section*{Acknowledgments}

We thank the Livo Team for their feedback and support throughout this project.

\appendix
\makeatletter
\setlength{\@fptop}{0pt}
\setlength{\@fpbot}{0pt plus 1fil}
\makeatother
\setcounter{table}{0}
\renewcommand{\thetable}{A\arabic{table}}
\setcounter{figure}{0}
\renewcommand{\thefigure}{A\arabic{figure}}
\section{State Slot Schema}
\label{app:slot-schema}

\paragraph{Schema organization and slot types.}
The schema organizes $\mathcal{M}_t$ into 40 core slot dimensions across seven semantic families (Tables~\ref{tab:slot-schema}--\ref{tab:slot-schema-context}): (1) \textit{Participant states} (\textit{user}, \textit{avatar}, and \textit{role} families) distinguish the real user from their role-play persona while tracking the character's dispositions and goals; (2) \textit{Dyadic dynamics} (\textit{relationship} family) capture evolving trust, boundaries, milestones, and unresolved open loops; and (3) \textit{Shared context} (\textit{emotional-event}, \textit{collective}, and \textit{storyworld} families) maintain pivotal shared memories, group alliances, and world rules. Each dimension is instantiated as either a \textit{text} slot (cohesive narrative syntheses, e.g., how joint interview preparation fostered mutual trust) or a \textit{structured} slot (discrete keyed records, e.g., specific interview anxieties and dates). A slot identifier combines a group prefix and a field, such as \texttt{user.} and \texttt{preference.life} in \texttt{user.preference.life}. Each entry retains attribution, its temporal scope, and provenance pointers to $\mathcal{E}_t$.

\paragraph{Epistemic grounding and information boundaries.}
\label{app:information-boundaries}
To ensure realistic multi-session and multi-party interactions, all stages---consolidation, retrieval, source-link resolution, and generation---enforce strict epistemic constraints: (1) \textit{causality}, restricting access to records strictly prior to turn $t$; (2) \textit{visibility}, restricting reads from $\mathcal{E}_t$ and $\mathcal{M}_t$ to records accessible to the responding character; (3) \textit{attribution}, separating the real-world user from their assumed fictional avatar; and (4) \textit{epistemic calibration}, preserving temporal scopes and distinguishing recorded observations from tentative interpretations. Current explicit requests constrain the response. Turn-level reasoning guides the current reply without being written back as factual evidence.

\paragraph{Reasoner input scope.}
At query time, \textit{Mind} reads the accessible contents of all 40 core slot dimensions alongside the current dialogue and initial vector candidates. The schema fixes the number of dimensions, not a token budget; input length depends on stored content and retrieved candidates.

\paragraph{Reasoner interface and action space.}
\label{app:mind-action-space}
The input is $(u_t,c_t,p,\mathcal{M}_t,\mathcal{C}_t)$: the current utterance, local dialogue and scene context, character specification, accessible state entries with provenance, and initial episodic candidates. The output is $z_t=(\tau_t,\mathcal{S}_t,\mathcal{V}_t)$. The interpretation $\tau_t$ distinguishes evidence from tentative needs and response intentions; $\mathcal{S}_t$ selects supplied state-entry IDs, and $\mathcal{V}_t$ selects supplied candidate IDs. Either list may be empty. The retrieval pipeline, rather than the reasoner, resolves selected provenance links under the same temporal and visibility constraints.

For illustration, suppose the accessible states include \texttt{s1} (an active interview goal, linked to $e_3,e_4$) and \texttt{s2} (a general preference for solitude), while vector candidate \texttt{c1} contains an earlier solitude-related episode. Given a current expression of low mood, the following JSON schematically represents the decision tuple:
\begin{quote}
\small
\begin{verbatim}
{
  "tau": "Tentatively ask about interview.",
  "S": ["s1"],
  "V": []
}
\end{verbatim}
\end{quote}
The retrieved evidence is then $\mathcal{L}_t=\{e_3,e_4\}$ despite the empty vector selection. These episodes support a tentative check-in.

\begin{table}[!tp]
\centering
\fontsize{10}{12}\selectfont
\begin{tabularx}{\columnwidth}{@{}>{\raggedright\arraybackslash}p{0.52\columnwidth} >{\raggedright\arraybackslash}X@{}}
\toprule
\textbf{Field} & \textbf{Content}\\\midrule
\multicolumn{2}{@{}l@{}}{\textbf{User (12)}\quad Prefix: \texttt{user.\allowbreak }}\\
\texttt{reality.\allowbreak context} & Real-world background\\
\texttt{preference.\allowbreak life} & Lifestyle preferences\\
\texttt{history.\allowbreak significant} & Significant life history\\
\texttt{goal} & Goals and progress\\
\texttt{preference.\allowbreak interaction} & Interaction preferences\\
\texttt{preference.\allowbreak experience\_\allowbreak focus} & Relationship/story focus\\
\texttt{preference.\allowbreak story} & Story preferences\\
\texttt{preference.\allowbreak relationship} & Desired relationship\\
\texttt{preference.\allowbreak tension} & Relational tension preferences\\
\texttt{psychology.\allowbreak pattern} & Psychological patterns\\
\texttt{support.\allowbreak pattern} & Effective support\\
\texttt{self.\allowbreak view} & Self-understanding\\
\midrule
\multicolumn{2}{@{}l@{}}{\textbf{Avatar (3)}\quad Prefix: \texttt{avatar.\allowbreak }}\\
\texttt{identity} & Fictional identity\\
\texttt{capability} & Fictional capabilities\\
\texttt{goal} & Fictional goals\\
\midrule
\multicolumn{2}{@{}l@{}}{\textbf{Role (10)}\quad Prefix: \texttt{role.\allowbreak }}\\
\texttt{identity} & Identity and social position\\
\texttt{background} & Formative background\\
\texttt{values\_\allowbreak principles} & Values and principles\\
\texttt{inner\_\allowbreak landscape} & Motives and vulnerabilities\\
\texttt{goal} & Goals and priorities\\
\texttt{competence} & Skills and limitations\\
\texttt{life\_\allowbreak pattern} & Habits and routines\\
\texttt{expression\_\allowbreak style} & Language and expression\\
\texttt{affection\_\allowbreak style} & Expressions of affection\\
\texttt{social\_\allowbreak network} & Social relationships\\
\bottomrule
\end{tabularx}
\caption{Core slot dimensions for the user, avatar, and role (25 total). Field names are relative to the displayed prefix.}
\label{tab:slot-schema}
\end{table}

\begin{table}[!tp]
\centering
\fontsize{10}{12}\selectfont
\begin{tabularx}{\columnwidth}{@{}>{\raggedright\arraybackslash}p{0.52\columnwidth} >{\raggedright\arraybackslash}X@{}}
\toprule
\textbf{Field} & \textbf{Content}\\\midrule
\multicolumn{2}{@{}l@{}}{\textbf{Relationship (10)}\quad Prefix: \texttt{relationship.\allowbreak }}\\
\texttt{position} & Relationship definition\\
\texttt{stage} & Developmental stage\\
\texttt{state} & Current relational state\\
\texttt{framework} & Shared rules and boundaries\\
\texttt{turning\_\allowbreak point} & Relationship-changing events\\
\texttt{milestone} & Shared milestones\\
\texttt{significant\_\allowbreak contribution} & Meaningful contributions\\
\texttt{gift} & Shared gifts and keepsakes\\
\texttt{emotional\_\allowbreak residue} & Lasting emotional effects\\
\texttt{open\_\allowbreak loop} & Unresolved shared matters\\
\midrule
\multicolumn{2}{@{}l@{}}{\textbf{Emotional event (1)}\quad Prefix: \texttt{memory.\allowbreak }}\\
\texttt{emotional\_\allowbreak event} & Emotionally significant events\\
\midrule
\multicolumn{2}{@{}l@{}}{\textbf{Collective (1)}\quad Prefix: \texttt{collective.\allowbreak }}\\
\texttt{relationship} & Group relations and alliances\\
\midrule
\multicolumn{2}{@{}l@{}}{\textbf{Storyworld (3)}\quad Prefix: \texttt{storyworld.\allowbreak }}\\
\texttt{state} & World facts and rules\\
\texttt{history} & Causal event history\\
\texttt{open\_\allowbreak loop} & Unresolved narrative threads\\
\bottomrule
\end{tabularx}
\caption{Core slot dimensions for relationships and shared context (15 total). Together with Table~\ref{tab:slot-schema}, these enumerate all 40 dimensions.}
\label{tab:slot-schema-context}
\end{table}

\paragraph{Runtime and reasoning.} Runtime context supplies the active role, scene summary, and participants. These inputs are distinct from the 40 persistent slot dimensions. Turn-specific \textit{Mind} output is used for reasoning and selection rather than stored as independent historical evidence.

\paragraph{Attribution and revision.} Real-user context, life preferences, significant personal history, and goals use evidence about the real user. Avatar identity, capabilities, and goals concern the fictional participant. Interaction preferences and psychological patterns may draw on explicit user statements and recurring choices, while preserving their evidence basis. Event-oriented entries retain historical scope; open-loop entries can be closed, cancelled, or invalidated as subsequent observations resolve them. Source links connect entries derived from a shared event to the same underlying evidence.

\section{Implementation and Training Details}
\label{app:implementation}
\subsection{Training Procedure}
\label{app:training-procedure}
\paragraph{Sample representation.}
Each supervised example pairs the reasoner input $x=(u,c,p,\mathcal{M},\mathcal{C})$ with the reference output $z^*=(\tau^*,\mathcal{S}^*,\mathcal{V}^*)$. State-selection targets refer to entries in the supplied state store, and episodic-selection targets refer to the supplied vector candidates. The reference \textit{Thinking} trace and both selection targets are generated by Claude Opus 5.5 and manually verified. Source episodes reached through state provenance are resolved after selection and are excluded from $\mathcal{V}^*$ (Appendix~\ref{app:reward-spec}). This representation couples the reference \textit{Thinking} trace with the two memory-selection targets in a single supervised output, including empty selections where applicable.

\paragraph{Optimization procedure.}
SFT initializes \textit{Mind} to produce Thinking and selected memories. GRPO then refines the SFT policy on the same 2,000 training inputs using downstream response rewards. For each sampled decision, the retrieval pipeline resolves selected state entries' source links and assembles their evidence with selected vector candidates before the fixed generator produces a reply. An LLM judge evaluates that reply for the response-quality reward; format validity and selection accuracy provide additional feedback, with a KL penalty to the SFT reference. Consolidation and vector retrieval remain fixed. We evaluate SFT and SFT+GRPO checkpoints separately to measure the additional benefit of GRPO. Data sources and partitioning are given in Section~\ref{sec:experimental-setup}, and hyperparameters are listed in Appendix~\ref{app:experimental-config}.

\subsection{Reward Specification}
\label{app:reward-spec}
This subsection details the composite reward formulation introduced in Section~\ref{sec:training-mind}, including ID-matching conventions and the response-quality judge rubric.

\paragraph{Format validity.} Let $g(z)\in\{0,1\}$ indicate whether \textit{Mind} output is parseable, contains the required fields, and uses admissible references: state-entry IDs must belong to the input slot store, and episodic IDs must belong to the vector candidates. Invalid outputs receive zero total reward.

\paragraph{Selection accuracy.} For annotated training targets $(\mathcal{S}^*,\mathcal{V}^*)$, we define
\[
a(x,z)=\tfrac12 F_1(\mathcal{S},\mathcal{S}^*)
       +\tfrac12 F_1(\mathcal{V},\mathcal{V}^*),
\]
where $F_1(A,B)=2|A\cap B|/(|A|+|B|)$ for a nonzero denominator. We set $F_1(\varnothing,\varnothing)=1$; if exactly one set is empty, its score is zero. Matching operates on exact entry IDs at the annotated state-entry and candidate-evidence granularity. Source-linked episodes resolved deterministically after slot selection are excluded from the candidate-selection target $\mathcal{V}^*$.

\paragraph{Response quality.} The reward judge assigns an integer score $s(x,y)\in\{0,1,2,3,4\}$, normalized as $q(x,y)=s(x,y)/4$ according to the rubric in Table~\ref{tab:reward-rubric}. Its evaluation context comprises the character specification, current conversation, and historical evidence available at the sample cutoff (excluding future turns). The response-quality judge uses the following instruction:
\begin{quote}\raggedright
Evaluate the reply using historical grounding, character and relationship consistency, and appropriateness to the current interaction. Distinguish supported statements from uncertain interpretations. Do not reward length, explicit memory mentions, or greater intimacy by themselves. Apply the 0--4 anchors in Table~\ref{tab:reward-rubric}. Return an integer score and a short justification tied to the supplied context. Do not use future dialogue or benchmark test annotations.
\end{quote}

\begin{table}[!htbp]
\centering
\begin{tabularx}{\columnwidth}{@{}r X@{}}
\toprule
\textbf{Score} & \textbf{Scoring anchor}\\\midrule
0 & Seriously contradicts evidence, character constraints, or explicit requests.\\
1 & Clearly inappropriate or relies on a consequential unsupported assumption.\\
2 & Broadly reasonable, but generic or misses consequential context.\\
3 & Appropriately uses history and fits the current interaction.\\
4 & Integrates relevant history and current context naturally, with uncertainty calibrated to the evidence.\\\bottomrule
\end{tabularx}
\caption{Training-reward rubric for response quality.}
\label{tab:reward-rubric}
\end{table}

\subsection{Experimental Configuration}
\label{app:experimental-config}
Table~\ref{tab:experimental-config} summarizes the model, training, and inference hyperparameters.

\begin{table}[!htbp]
\centering\small
\renewcommand{\arraystretch}{0.99}
\begin{tabularx}{\columnwidth}{@{}>{\raggedright\arraybackslash}p{0.38\columnwidth} >{\raggedright\arraybackslash}X@{}}
\toprule
\textbf{Setting} & \textbf{Value}\\\midrule
Memory schema & 40 core dimensions (Tables~\ref{tab:slot-schema}--\ref{tab:slot-schema-context})\\
\textit{Mind} state input & All 40 state dimensions in full\\
\textit{Mind} backbones & Qwen3.5 (0.8B, 2B)\\
Evaluated checkpoints & SFT and SFT+GRPO\\
General-purpose reasoner & \texttt{deepseek-v4-flash}\\
Training stages & SFT, followed by GRPO\\
SFT examples & 2,000\\
GRPO inputs & Same 2,000 inputs as SFT\\
SFT learning rate & $10^{-5}$\\
SFT global batch size & 32\\
SFT batch per device & 8\\
SFT epochs & 3\\
GRPO group size $G$ & 8\\
GRPO learning rate & $5\times10^{-6}$\\
GRPO training steps & 200 optimizer steps\\
KL coefficient $\beta$ & 0.04\\
Reward weights & $\lambda_{\mathrm{fmt}}=0.10$, $\lambda_{\mathrm{acc}}=0.30$, $\lambda_{\mathrm{qual}}=0.60$\\
Training generator & \texttt{deepseek-v4-pro}\\
Reward judge & \texttt{claude-sonnet-5.5}\\
Embedding model & \texttt{text-embedding-v4}\\\midrule
\multicolumn{2}{@{}l}{\textit{MemReal-Bench evaluation}}\\
Response generator & \texttt{deepseek-v4-pro}\\
Judge & \texttt{deepseek-v4-flash}\\
Consolidator & \texttt{deepseek-v4-flash}; 10 new memory records trigger an update; 20 recent dialogue turns per update\\
\textbf{MindMem} initial vector candidates & 20\\
Baseline recall entries & Up to 16\\
\textbf{MindMem} final evidence records & Up to 16, assembled from vector-selected and slot-linked evidence\\
\textbf{MindMem} active slots & Query-dependent; provided alongside the evidence records\\
Temperature & 0\\
Response token limit & 2,048\\
SMR judge decoding & 4,096 tokens; \texttt{json\_object}\\
DQA judge decoding & 5 tokens; yes / no\ (Table~\ref{tab:prompt-decoding})\\
SMR recent dialogue & Last 10 visible turns (except \textit{NoMem})\\
Character context & Public identity\\
Scene context & One-line description\\
Selector prompt & State-slot and candidate episode selection with \textit{Thinking}\\
Active recall & One \textit{Mind} call generates \textit{Thinking} and selects state slots and candidate episodes\\
w/o active recall & Remove the active-recall pathway. Retain up to 16 vector-retrieved dialogue spans\\\midrule
\multicolumn{2}{@{}l}{\textit{Standard benchmarks}}\\
Protocol & Mem0 \texttt{memory-benchmarks} runner, memory client swapped per system\\
Answerer / judge & \texttt{deepseek-v4-flash} (temperature 0)\\
Retrieval request & top-200 candidates, time-ordered\\
Question coverage & LoCoMo 1,540 (adversarial excluded); LongMemEval-S 500\\
\bottomrule
\end{tabularx}
\caption{Training and evaluation settings for \textbf{MemReal-Bench} and the standard memory benchmarks.}
\label{tab:experimental-config}
\end{table}

\paragraph{Recall configuration.} The active-slot count refers to selected state entries and varies by query; \textit{Mind} reads all 40 state dimensions. Selected slots' provenance links are resolved into source evidence and assembled jointly with selected vector-retrieval evidence (Section~\ref{sec:active-recall}). In \textbf{MemReal-Bench}, the final evidence set $\mathcal{D}_t$ contains at most 16 records across these two routes, with slot-linked evidence included in this cap. Selected state slots and \textit{Thinking} guidance are supplied separately to the generator. Baselines and \textit{w/o active recall} provide up to 16 vector-recalled entries; \textbf{Full} combines vector-selected and slot-linked episodes within the same cap, alongside selected states and \textit{Thinking}. The cap is defined by evidence-record count; record lengths and total generator input tokens vary across systems.

\paragraph{Standard benchmarks.} LoCoMo and LongMemEval-S follow Mem0's open-source \texttt{memory-benchmarks} protocol (Section~\ref{sec:experimental-setup}), with only the memory client replaced per system: \texttt{deepseek-v4-flash} as answerer and judge, \texttt{text-embedding-v4} embeddings, and top-200 retrieval, for the 1{,}540-question non-adversarial LoCoMo split and the 500-question LongMemEval-S benchmark. Each question is ingested into an isolated memory space, sessions are written in date order with their dates attached, and retrieved candidates are time-ordered before answering.

\section{MemReal-Bench Construction and Evaluation Protocol}
\label{app:benchmark-protocol}
This appendix details the construction pipeline, evaluation dimensions, and quality-control protocols for \textbf{MemReal-Bench} (Section~\ref{sec:benchmark}). Table~\ref{tab:corpus-coverage} summarizes the evaluated dialogue and scored instances.

\begin{table}[!htbp]
\centering\small
\renewcommand{\arraystretch}{1.12}
\begin{tabularx}{\columnwidth}{@{}X r@{}}
\toprule
\textbf{Statistic} & \textbf{Value}\\\midrule
Evaluated dialogue turns & 4,617\\
Dialogue tokens (\texttt{cl100k\_base}) & 443,994\\
Situated Memory Realization (SMR) probes & 310\\
Scored DQA questions & 511\\\bottomrule
\end{tabularx}
\caption{Statistics of the evaluated \textbf{MemReal-Bench} dialogue and scored instances.}
\label{tab:corpus-coverage}
\end{table}

\subsection{Evaluation Dimensions}
\label{app:benchmark-dimensions}
Table~\ref{tab:scoring} defines the five SMR dimensions introduced in Section~\ref{sec:benchmark-metrics}.

\begin{table}[!htbp]
\centering\small
\renewcommand{\arraystretch}{1.08}
\begin{tabularx}{\columnwidth}{@{}>{\raggedright\arraybackslash}p{0.33\columnwidth} >{\raggedright\arraybackslash}X@{}}
\toprule
\textbf{SMR dimension} & \textbf{History-specific realization criterion}\\\midrule
Factual Realization (FR) & Uses a past fact or event to inform the current response or revise a plan.\\\addlinespace[2pt]
World Consistency (WC) & Lets established world rules and states constrain proposed actions.\\\addlinespace[2pt]
Relational Realization (RR) & Adapts familiarity, interpersonal distance, or commitments to the relationship supported by history.\\\addlinespace[2pt]
Role Realization (RoR) & Enacts the character's established principles, dispositions, or behavioral patterns.\\\addlinespace[2pt]
Personalization (PI) & Tailors a response or choice to the user's historically established preferences or patterns.\\\bottomrule
\end{tabularx}
\caption{Situated Memory Realization (SMR) dimensions in \textbf{MemReal-Bench}. All probes require target-memory utilization not inferable from the current turn alone.}
\label{tab:scoring}
\end{table}

\paragraph{Aggregation and denominators.} Let $\mathcal{I}$ denote the scored probe--dimension pairs, with one primary dimension per probe and $|\mathcal{I}|=310$, and let $s_{i,d}\in\{0,1\}$ be the binary verdict from the identity-blind LLM judge; overall SMR is the micro-average $\operatorname{SMR}=\frac{100}{310}\sum_{(i,d)\in\mathcal{I}}s_{i,d}$. Per-dimension denominators are 143, 33, 47, 52, and 35 for FR, WC, RR, RoR, and PI, respectively. DQA answers are scored against annotated answer points, permitting semantic paraphrases, and overall accuracy is micro-averaged over 511 questions. Because SMR and DQA evaluate complementary aspects over distinct denominators, they are reported alongside each other rather than merged.

\paragraph{Paired diagnostic aggregation.} Let $P$ be the evaluated FR/WC probes with linked DQA questions, and let $m_i\geq1$ be the number of questions linked to probe $i$. For system $s$, let $a_{sij}\in\{0,1\}$ indicate correctness on question $j$ of probe $i$, and let $y_{si}\in\{0,1\}$ be its SMR verdict. The two component rates are
\[
\begin{aligned}
\mathrm{DQA}_{s,P}
&=100\,\frac{\sum_{i\in P}\sum_{j=1}^{m_i}a_{sij}}
{\sum_{i\in P}m_i},\\
\mathrm{SMR}_{s,P}
&=100\,\frac{\sum_{i\in P}y_{si}}{|P|}.
\end{aligned}
\]
The reported Gap is
\[
\mathrm{Gap}_s=\mathrm{DQA}_{s,P}-\mathrm{SMR}_{s,P},
\]
in percentage points. Gap uses all 176 linked FR/WC probes (143 FR and 33 WC) and their 511 DQA questions. DQA is micro-averaged by question and SMR by probe.

For the conditional realization analysis, define
\[
\begin{aligned}
K_s&=\{i\in P:a_{sij}=1\ \text{for all }j\},\\
n_s&=|K_s|,\\
\mathrm{Real}_s&=100\,\frac{\sum_{i\in K_s}y_{si}}{n_s}.
\end{aligned}
\]
Real. uses each system's subset $K_s$, consisting of probes for which it answers every linked DQA question correctly. We calculate Gap from unrounded component rates. At a fixed SMR, lower DQA accuracy yields a smaller Gap.

\subsection{Construction Pipeline and Quality Control}
\label{app:benchmark-construction}
\textbf{MemReal-Bench} is constructed via a three-stage construction pipeline:
\begin{enumerate}
\item \textbf{Seed Distillation:} We distill user personas, interaction habits, and relational progression trajectories from a month-long source collection of real-user and character interactions. Raw chat logs are excluded from downstream dialogue-generation inputs and benchmark histories.
\item \textbf{Longitudinal History Synthesis:} Distilled seeds are paired with character and storyworld specifications to synthesize multi-session dialogue mixing dyadic and group interactions. Every turn is indexed with participant identities, session/turn coordinates, and character-specific visibility scopes.
\item \textbf{Evidence-First Probe Construction:} We extract atomic memory records (propositions, verbatim quotes, source turns, and temporal/visibility scopes) directly from the realized transcripts. Each probe includes a natural trigger and an SMR rubric. FR/WC probes additionally include explicit DQA questions.
\end{enumerate}

\paragraph{Temporal and visibility verification.}
The split unit is the source-user seed: histories sharing a seed belong to the same partition. For each probe $i$ with cutoff $b_i$ and responding character $r_i$, the verification protocol checks that: (1) all target evidence occurs at or before $b_i$ within scenes visible to $r_i$; (2) superseded states are marked with validity intervals so models are evaluated on the latest valid state; and (3) natural triggers cue the situational context without lexically leaking the target memory.

\begin{table*}[!t]
\centering\small
\renewcommand{\arraystretch}{1.08}
\begin{tabular*}{\textwidth}{@{\extracolsep{\fill}}l r rrrr@{}}
\toprule
& \textbf{v4-flash} & \multicolumn{2}{c}{\textbf{Qwen3.5-0.8B}} & \multicolumn{2}{c}{\textbf{Qwen3.5-2B}}\\
\cmidrule(lr){3-4}\cmidrule(l){5-6}
\textbf{Metric} & \textbf{General-purpose} & \textbf{SFT} & \textbf{SFT+GRPO} & \textbf{SFT} & \textbf{SFT+GRPO}\\\midrule
FR & 44.76 & 56.64 & 58.04 & 59.44 & \textbf{61.54}\\
WC & 57.58 & 54.55 & 57.58 & 60.61 & \textbf{66.67}\\
RR & 40.43 & 36.17 & 38.30 & 51.06 & \textbf{55.32}\\
RoR & 38.46 & 46.15 & 50.00 & 50.00 & \textbf{61.54}\\
PI & 22.86 & 31.43 & 34.29 & 37.14 & \textbf{42.86}\\\midrule
SMR (all) & 41.94 & 48.71 & 50.97 & 54.19 & \textbf{59.03}\\
DQA (all) & 80.43 & 81.21 & 81.02 & 82.00 & \textbf{82.58}\\\midrule
Latency (s) & 3.5 & \multicolumn{2}{c}{\textbf{1.1}} & \multicolumn{2}{c}{1.7}\\\bottomrule
\end{tabular*}
\caption{\textit{Mind} selector and training-stage comparison on \textbf{MemReal-Bench}. SMR and DQA are percentages over the same 310 probes and 511 questions, respectively. All configurations share active recall, the response generator, and the judge; SFT+GRPO-2B is the \textbf{MemReal-Bench} configuration of Table~\ref{tab:main}. Latency is the median wall-clock time from selector invocation to completion of its output. Values are rounded to one decimal place, with one latency value per model size across SFT and SFT+GRPO. Bold marks the best available score per row.}
\label{tab:scale}
\end{table*}

\section{Ablation and Diagnostic Analysis Details}
\label{app:analysis-protocol}
This appendix provides the experimental protocols and diagnostic criteria supporting the ablation study (Section~\ref{sec:ablations}) and error analysis (Section~\ref{sec:know-use}).

\subsection{Mind Selector Comparison}

Table~\ref{tab:scale} compares the general-purpose \texttt{deepseek-v4-flash} reasoner, used in the \textit{w/o specialized Mind} condition, with specialized Qwen3.5 reasoners trained with SFT or SFT+GRPO at 0.8B and 2B. All configurations use the same active recall interface. Comparisons between SFT and SFT+GRPO at each model size assess the additional benefit of GRPO, while comparisons with \texttt{deepseek-v4-flash} assess the specialized \textit{Mind} relative to a general-purpose replacement.

\paragraph{Effect of GRPO.}
Compared with SFT, GRPO improves SMR by 2.26 and 4.84 percentage points at 0.8B and 2B, respectively (across all five dimensions), while keeping DQA essentially stable (changes below one point: $-0.19$ and $+0.58$). At one-decimal precision, SFT and SFT+GRPO share median selection latencies of 1.1 seconds at 0.8B and 1.7 seconds at 2B. These matched-backbone comparisons show higher observed SMR after GRPO, with small changes in DQA and comparable median selection latency under the evaluated settings.

\subsection{Diagnostic Taxonomy for Failure Localization}
\label{app:failure-taxonomy}

The audit in Section~\ref{sec:know-use} uses two categories, defined by whether the stored target memory reaches the final generator input.
\begin{enumerate}
\item \textbf{Recall Miss:} The target memory is stored in the memory system but absent from the final generator input.
\item \textbf{Realization Failure:} The target memory is present in the final generator input, but the response fails to realize it appropriately.
\end{enumerate}
These categories localize failures at the recall--realization boundary according to whether the target memory reaches the final generator input.

\subsection{Paired System Comparison}
\label{app:paired-system-comparison}
Using the archived judgments behind Table~\ref{tab:main}, we pair the binary verdicts for \textbf{MindMem} and A-Mem by probe ID for SMR and question ID for DQA. We enumerate paired permutations by swapping the two system labels jointly within each cutoff session. The two-sided test statistic is the absolute difference between the micro-averaged scores. There are 42 cutoff sessions with scored SMR probes and 39 with scored DQA questions. With blocks of four consecutive historical sessions, the corresponding $p$-values are 0.00098 for SMR and 0.0244 for DQA.

\onecolumn

\providecommand{\revadd}[1]{#1}

\newenvironment{prompttable}
  {\par\addvspace{5pt}\noindent\begin{minipage}{\textwidth}\centering
   \captionsetup{type=table,skip=4pt}}
  {\end{minipage}\par\addvspace{5pt}}

\newcommand{\promptbox}[2]{%
  \centering
  \setlength{\tabcolsep}{6pt}%
  \renewcommand{\arraystretch}{1.1}%
  \begin{tabularx}{\textwidth}{|X|}
  \hline
  \textbf{#1}\\
  \hline
  \begin{minipage}{\linewidth}
  \vspace{1pt}
  \VerbatimInput[fontsize=\footnotesize,baselinestretch=0.9,breaklines=true,breakanywhere=true,breaksymbolleft={},breaksymbolright={},tabsize=4]{#2}
  \vspace{1pt}
  \end{minipage}\\
  \hline
  \end{tabularx}}

\section{Evaluation Prompt Details}
\label{app:evaluation-prompts}
This appendix presents the active-recall, response-generation, and evaluation prompts \revadd{for \textbf{MemReal-Bench} (Tables~\ref{tab:prompt-active-system-1}--\ref{tab:prompt-dqa-judge-user}) and their decoding settings (Table~\ref{tab:prompt-decoding})}. We provide English versions of the Chinese-language prompt templates; the direct-QA judge prompts were originally in English. Braced expressions denote runtime substitutions. All displayed JSON keys and code identifiers retain their original names.

\subsection{Active Recall}
\label{app:prompt-active-recall}
The system prompt defines the selection policy and output format; the user prompt supplies the query-specific inputs (Tables~\ref{tab:prompt-active-system-1}--\ref{tab:prompt-active-user}). \revadd{The system prompt spans two tables, split before its \texttt{[selected\_slots]} section.} The selection cap counts content items. \texttt{[Available Slots]} contains both state entries and vector-retrieved episodic candidates. The \texttt{selected\_slots} field returns selections from both groups ($\mathcal{S}_t$ and $\mathcal{V}_t$, respectively).

\begin{prompttable}
\promptbox{ACTIVE\_RECALL\_SYSTEM\_PROMPT (Part 1 of 2)}{doc/prompt_templates/active-system-1.txt}
\caption{System prompt for active recall: intent interpretation, memory selection, and memory-use guidance \revadd{(Part 1 of 2; continued in Table~\ref{tab:prompt-active-system-2})}.}
\label{tab:prompt-active-system-1}
\end{prompttable}

\begin{prompttable}
\promptbox{ACTIVE\_RECALL\_SYSTEM\_PROMPT (Part 2 of 2, continued)}{doc/prompt_templates/active-system-2.txt}
\caption{System prompt for active recall\revadd{, continued from Table~\ref{tab:prompt-active-system-1} (Part 2 of 2): selection priorities and output format}.}
\label{tab:prompt-active-system-2}
\end{prompttable}

\begin{prompttable}
\promptbox{ACTIVE\_RECALL\_USER\_PROMPT}{doc/prompt_templates/active-user.txt}
\caption{User prompt for active recall, supplying the current context, selection constraints, and candidate memories.}
\label{tab:prompt-active-user}
\end{prompttable}

\newpage
\subsection{Response Generation}
\label{app:prompt-response-generation}
Tables~\ref{tab:prompt-response-system}--\ref{tab:prompt-dqa-user} give the response templates. The memory-text variable contains the complete recalled context supplied to the generator. The scene variable contains the last 10 visible turns and is empty for NoMem. The basic-information line is omitted when that variable is empty; the scene section is omitted when the scene variable is empty. DQA calls use an empty scene context.

\begin{prompttable}
\promptbox{NATURAL\_RESPONDER\_SYSTEM\_PROMPT}{doc/prompt_templates/response-system.txt}
\caption{System prompt for situated character responses.}
\label{tab:prompt-response-system}
\end{prompttable}

\begin{prompttable}
\promptbox{NATURAL\_RESPONDER\_USER\_PROMPT}{doc/prompt_templates/response-user.txt}
\caption{User prompt for situated responses, supplying relevant history, local context, and the current utterance.}
\label{tab:prompt-response-user}
\end{prompttable}

\begin{prompttable}
\promptbox{DIRECT\_QA\_RESPONDER\_SYSTEM\_PROMPT}{doc/prompt_templates/dqa-system.txt}
\caption{System prompt for direct question answering from recalled history.}
\label{tab:prompt-dqa-system}
\end{prompttable}

\begin{prompttable}
\promptbox{DIRECT\_QA\_RESPONDER\_USER\_PROMPT}{doc/prompt_templates/dqa-user.txt}
\caption{User prompt for direct question answering.}
\label{tab:prompt-dqa-user}
\end{prompttable}

\subsection{Automated Evaluation}
\label{app:prompt-automated-evaluation}
Tables~\ref{tab:prompt-smr-judge-system}--\ref{tab:prompt-dqa-judge-user} give the judge templates. The private gold object contains the applicable metrics, target memories, and per-metric rubrics. In direct QA, the expected answer joins the annotated answer points, falling back to the question if absent; the optional paraphrase line supplies acceptable alternatives.
\par\nopagebreak[4]

\begin{prompttable}
\promptbox{NATURAL\_JUDGE\_SYSTEM\_PROMPT}{doc/prompt_templates/smr-judge-system.txt}
\caption{System prompt for evaluating situated memory realization.}
\label{tab:prompt-smr-judge-system}
\end{prompttable}

\begin{prompttable}
\promptbox{NATURAL\_JUDGE\_USER\_PROMPT}{doc/prompt_templates/smr-judge-user.txt}
\caption{User prompt for evaluating situated memory realization against private target annotations.}
\label{tab:prompt-smr-judge-user}
\end{prompttable}

\begin{prompttable}
\promptbox{DIRECT\_QA\_JUDGE\_SYSTEM\_PROMPT}{doc/prompt_templates/dqa-judge-system.txt}
\caption{System prompt for binary direct-answer evaluation.}
\label{tab:prompt-dqa-judge-system}
\end{prompttable}

\begin{prompttable}
\promptbox{DIRECT\_QA\_JUDGE\_USER\_PROMPT}{doc/prompt_templates/dqa-judge-user.txt}
\caption{User prompt for comparing a direct answer with its expected facts.}
\label{tab:prompt-dqa-judge-user}

\end{prompttable}

\subsection{Decoding Parameters}
\label{app:prompt-decoding-parameters}
Table~\ref{tab:prompt-decoding} summarizes decoding settings for the \textbf{MemReal-Bench} prompt roles.

\begin{prompttable}
{\centering\small
\setlength{\tabcolsep}{5pt}
\begin{tabularx}{\textwidth}{@{}lXrrl@{}}
\toprule
\textbf{Role} & \textbf{Model} & \textbf{Temperature} & \textbf{Max.\ tokens} & \textbf{Output constraint}\\
\midrule
Active recall & Qwen3.5-2B (SFT+GRPO; self-hosted) & 0 & 1536 & JSON\\
Response (SMR / DQA) & \texttt{deepseek-v4-pro} & 0 & 2048 & ---\\
SMR judge & \texttt{deepseek-v4-flash} & 0 & 4096 & \texttt{json\_object}\\
DQA judge & \texttt{deepseek-v4-flash} & 0 & 5 & Yes / no\\
\bottomrule
\end{tabularx}
\caption{Decoding settings for active recall, response generation, and automated evaluation on \textbf{MemReal-Bench}.}
\label{tab:prompt-decoding}\par}
\end{prompttable}

\end{document}